\documentclass[letterpaper, 10 pt, conference]{ieeeconf}  

\IEEEoverridecommandlockouts                              

\usepackage{svg}
\usepackage{relsize}
\usepackage{graphics} 
\usepackage{epsfig} 
\usepackage[export]{adjustbox}
\usepackage{amsmath} 
\usepackage{amssymb}  
\usepackage{textgreek}
\usepackage{cite}
\usepackage{subcaption}

\title{\LARGE \bf
Human-Robot Skill Transfer with Enhanced Compliance via Dynamic Movement Primitives
}

\author{Jayden Hong$^{1}$, Zengjie Zhang$^{2}$, Amir M. Soufi Enayati$^{1}$, Homayoun Najjaran$^{1}$
\thanks{This work was supported by Kinova\textregistered~Inc. and Natural Sciences and Engineering Research Council (NSERC) Canada under the Grant CRDPJ 543881-19.}
\thanks{$^{1}$Jayden Hong, Amir M. Soufi Enayati and Homayoun Najjaran are with the Faculty of Engineering and Computer Science, University of Victoria, Victoria BC, Canada
        {\tt\small \{jaydenh, amsoufi, najjaran\}@uvic.ca}}%
\thanks{$^{2}$Zengjie Zhang is with the Department of Electrical Engineering, Eindhoven University of Technology, Eindhoven, Netherlands
        {\tt\small z.zhang3@tue.nl}}%
}
\usepackage{multirow}

\begin{document}

\maketitle
\thispagestyle{empty}
\pagestyle{empty}

\begin{abstract}

Finding an efficient way to adapt robot trajectory is a priority to improve overall performance of robots.
One approach for trajectory planning is through transferring human-like skills to robots by Learning from Demonstrations (LfD).
The human demonstration is considered the target motion to mimic. However, human motion is typically optimal for human embodiment but not for robots because of the differences between human biomechanics and robot dynamics.
The Dynamic Movement Primitives (DMP) framework is a viable solution for this limitation of LfD, but it requires tuning the second-order dynamics in the formulation.
Our contribution is introducing a systematic method to extract the dynamic features from human demonstration to auto-tune the parameters in the DMP framework. 
In addition to its use with LfD, another utility of the proposed method is that it can readily be used in conjunction with Reinforcement Learning (RL) for robot training.
In this way, the extracted features facilitate the transfer of human skills by allowing the robot to explore the possible trajectories more efficiently and increasing robot compliance significantly.
We introduced a methodology to extract the dynamic features from multiple trajectories based on the optimization of human-likeness and similarity in the parametric space.
Our method was implemented into an actual human-robot setup to extract human dynamic features and used to regenerate the robot trajectories following both LfD and RL with DMP. It resulted in a stable performance of the robot, maintaining a high degree of human-likeness based on accumulated distance error as good as the best heuristic tuning.

\end{abstract}

\section{Introduction}

The recent adaptation of AI into robotics has enhanced the application of Factory Automation technology. While recent advances in computer vision technologies have made manufacturing processes increasingly automated, a human expert is often required to teach the initial trajectory to match the master vision image, which is time-consuming and often requires the entire production line stopped until the commissioning is finished. Slight changes in the object a robot should engage with in the shape, size or orientation can potentially interfere or collide with other components. To overcome these challenges, it is mandatory for robots to have the ability to adapt to the changes in their environments and generate an optimal trajectory by themselves. 

Learning from Demonstration (LfD) has been one of the earliest approaches to transferring human knowledge to robots to give skill in generating trajectories. In LfD, the human demonstration is assumed to be the ground truth for a robot to imitate. However, the earliest methodologies of LfD training the robot trajectory directly from a human motion had limitations that made them infeasible for several reasons. First, the differences in an embodiment between a human and a robot could result in unstable or inaccurate trajectories. Second, the regenerated trajectories were often task-specific and lacked generalizability to variations in the environment. In addition, the mechanical impedance of the motor structure limits the capability to produce a jerky movement, unlike the human muscle which can respond rapidly. Therefore, the robot must balance two conflicting goals simultaneously~\cite{Billard2013}: closely mimicking human behavior on the one hand, while maintaining its stability and safety on the other.

Introducing a dynamic system (DS) into the control model in LfD can help to balance these two goals of human-likeness and motion stability in LfD. By modelling complex nonlinear dynamics into a simple one-degree-of-freedom (DoF) spring damper dynamic system with perturbed force, a robot can follow the demonstrated trajectory only within the compliance of the DS. The DS prevents making noisy or jerky movements and allows to generalize of the motion simply by extrapolating the learned motion to a new situation. One of the successful frameworks is the Dynamic Movement Primitives (DMP)~\cite{shaal2012dmp,ijspeert2013dynamical}. The DMP is formulated as a point attractor to the goal position \textit{g}. The gains of the DMP system correspond to the coefficient of assumed DS i.e., stiffness $K$, damping $D$ and inertia $K$ of DS. In~\cite{calinon2010learning}, DMP showed the best performance in reproducing smooth trajectory as well as the least computational cost among the other LfD methods, but the trajectory reproduction was only be stabilized with fine-tuned gains. 

Despite the success of the LfD approaches, human biomechanics can cast doubts about using human demonstrations as the perfect ground truth. The human motor system has more degrees of freedom (DoF) than its structural DoF, leading to multiple muscles being involved in 1 DoF motion. The activations of muscles are optimized to minimize muscle endurance~\cite{Crowninshield1981}, energy expenditure or muscle fatigue~\cite{prilutsky2002optimization}, which is different from the single motor joint movement of a robot. Since the maximum forces and endurance of the muscles depend on posture during motion, Humans avoid elbow-up posture, preferring elbow-down to minimize the stress on muscles and tendons, even if it's a shorter trajectory. In contrast, robots can generate uniform torques within the range of motion, making Optimal solutions for robots may differ from those for humans.

Reinforcement Learning (RL) with DMP can be a decent way to find out a more flexible solution optimized for the robot even if the solution might not be the preferred motion for a human. The RL agent explores the policy complying with the boundary of the robot's dynamic constraints. Although the DMP was originally devised as an LfD methodology, it can be well adapted to RL by substituting forcing terms for the action of RL agent. ~\cite{Stulp2012MFDMPRL,Li2021RLDMP,Yuan2019DMPRLexo}. As the DMP framework in LfD maps the trajectory in the task space onto the forcing term in DMP space, DMP in RL defines the mapping between state (i.e. trajectory) and action (i.e. forcing term). Nevertheless, RL with DMP has shortcomings in that the human intention can be delivered to the agent only by its reward, losing the benefit of using DMP's ability to deliver the human intention.

In both LfD and RL with DMP framework, auto-tuning of $M, D, K$ parameters can help to increase the degree of automation and boost the advantage of DMP. However, finding a proper set of $M, D, K$ parameters has a more significant meaning for skill transfer. As $M, D, K$ are the assumed coefficient of imaginary DS to formulate DMP, they determine the spring-damping responses of the DS. An inappropriate set of $M, D, K$ may lead to unstable or unresponsive motions in both frameworks. In other words, the optimal $M, D, K$ in DMP represents the spring-damping characteristic demonstrated trajectories and delivers it to the regenerated trajectory, which can provide a more general context of the human skill other than a trajectory. These $M, D, K$ parameters are referred to as dynamic features in the remaining part of the paper.

 According to the survey of recent skill transfer learning methodologies for robots~\cite{Liu2020skilltransfer}, the transferable skill has been identified as the two categories: demonstrated behavior such as trajectories and forces, and the neurophysiological signals such as human muscle activation and stiffness from electromyography (EMG). The approaches to transfer the stiffness of the human motion using EMG signal ~\cite{lu2021constrained, yu2022human, bian2019extended} has shown the possibility that dynamic characteristics can deliver human skill successfully as well as the trajectory itself in a way that the robot remained robust stability despite the unexpected external perturbation. However, it requires additional devices to sense the signal, and still inherits the limitation of the demonstration that the activation is specifically optimized for human anatomy.

In this paper, we suggest an approach to extract the dynamic features from the human trajectory to use both in LfD and RL with DMP frameworks. The suggested extraction methodology compares the demonstrated trajectories in the meta-parametric space of $M, D, K$ to find out the optimal dynamic features where DMP can stably regenerate trajectories without losing the likeness to its original demonstration. The rest of the paper is organized as follows: Section 2 presents the reasoning and the methodology to define the objective function to optimize and extraction of the parameters from it; Section 3 presents the experimental setup for demonstration and validation; the simulation result for its adaptation to a robot is presented in section 4; section 5 concludes the paper with future works.

\section{Dynamic Feature Extraction}

\subsection{Dynamic Movement Primitives}

In DMP, the point attractor system is formulated as 
\begin{align}
\tau_y \ddot y \,&= \alpha ( \beta (g-y) - \dot y ) + f, \\
\tau_x \dot x \,&=  x, 
\end{align}
where $y$ is the position, $\dot y$ is the velocity, $\ddot y$ is the acceleration of the trajectory, and $\alpha, \beta$ are the gains of the dynamic system, $\tau_y$ is the temporal scaling term of the trajectory, $f = \frac{\sum{\psi_i w_i}}{\sum{\psi_i}}x(g-y_0)$ is the forcing term, $\psi_i$ are Gaussian basis functions (BFs), $w_i$ are the weights of the BFs, $x$ is state variable with $x(0)=1$ where $t=0$ to make the forcing term converge from 1 to 0, $\tau_x$ is the convergence gain for $x$, $(x_0-g)$ is the spatial scaling term of DMP. In the joint space, each joint trajectory generates separate DMP forcing terms. In the task space, the Cartesian trajectories generate three DMP forcing terms, one each in the x, y, and z-axis. In this paper, the trajectory $y, \dot y, \ddot y$ are discussed and analyzed in the task space excluding its orientation.

The DMP can also be represented as a spring-damper DS 
\begin{equation}
\begin{aligned}
&\tau_y \ddot y + \alpha \dot y + \alpha \beta (y-g) \\
=&M\ddot y + D\dot y + K(y-g)= f  
\end{aligned}
\end{equation}
where M is the inertia, D is the damping coefficient, and K is the stiffness of the DS. The forcing term $f$ is approximated by the summation of the weighted Gaussian bases with different distributions. Generally, locally weighted regression is used to train the weights, also known as the shape parameters. Once the DMP is trained, the weights can be reused to generate other trajectories similar to the original demonstration with different goals. In the typical DMP modelling, the internal gains $\alpha$, $\beta$, $\tau$ (or dynamic features $M,D,K$) are assumed to have specific constant values and used to achieve the weights $w_i$. 

Most of the DMP models with both LfD and RL assume the fine-tuned dynamic features $M,D,K$. However, the manual tuning of the dynamic features limits the DMP's ability to regenerate trajectories autonomously. One approach for parameter tuning is to assume a critically damped system~\cite{ijspeert2013dynamical} with the damping ratio:
\begin{align}
\zeta=\frac{D}{2\sqrt{KM}} = 1
\end{align}
While the critical damping condition guarantees the system stability, it may not reflect the desired spring-damping characteristic of the demonstrated motion. However, in situations where both the target forcing term and the dynamic features are unknown in the system, achieving one from another has been explored by researchers, but adapting both simultaneously remains a challenge. The methodology suggested by Y. Cohen et al.~\cite{cohen2021motion} to adapt DMP parameters in the parameter manifold can give a clue to the dynamic feature extraction. They constructed the DMP over the parameter manifold to calculate the task result (landing position) of softball throwing and performed component analysis (PCA) and locally linear embedding (LLE) to adapt the throwing angles and position  n of the robot, which successfully showed the successful throwing motion in the experiment. Although they only adapted the goal position of the motion, the idea of parameterizing an objective function in the meta-parametric space can be implemented for the dynamic feature extraction. 

\subsection{Optimization of Dynamic Features}

To achieve stable motion generation that reflects human intention, the objective function of dynamic features in LfD should consider two potentially conflicting goals: human-likeness and motion stability.

For the human-likeness term, the accumulated euclidean distance error between the demonstrated trajectory and the regenerated trajectory for each set of $M, D, K$ can be compared to find the optimal dynamic features where both trajectories have the same temporal and spatial sizes. At first, The state equation of regenerated trajectory can be formulated as 

\begin{equation}
\begin{aligned}
\ddot y_{new} =& -\frac{D}{M} \dot y_{demo} - \frac{K}{M}(y_{demo}-g_{demo})
\\&+ \frac{\sum{\psi_i w_i}(y_{demo}(0)-g_{demo})}{\sum{\psi_i} M}x ,\\
\end{aligned}
\end{equation}
\begin{align}\label{eq:ydnew}
\dot y_{new} =& \ddot y_{new} \Delta t ,
\end{align}
\begin{align}\label{eq:ynew}
y_{new} =& \dot y_{new} \Delta t    
\end{align}

where $y_{demo}(0)$ is the initial position and $g_{demo}$ is the goal position of the demonstrated trajectory, $y_{new}, \dot y_{new}, \ddot y_{new}$ is the regenerated trajectory from the demonstrated trajectory $y_{demo}, \dot y_{demo}, \ddot y_{demo}$. As M is equivalent to temporal scaling factor $\tau_y$, different M changes the temporal size of the DMP forcing term by changing the size of $\Delta t$ in Eq.~\eqref{eq:ydnew} and~\eqref{eq:ynew}. M should be set equal to the time duration $T_{demo}$ which scales the temporal size of the state variables of the regenerated trajectory, in order to enable the comparison of the trajectories in the same unit time duration. It is also important to note that the optimal $D$ and $K$ depend on the desired duration of the motion or the time scaling factor $M$, and that $D$ and $K$ should increase linearly with $M$. Therefore, it is concluded that the extracted dynamic features are not a single set but the ratios of $M,D,K$:
\begin{gather}
D_M=\frac{D}{M},\text{ } K_M=\frac{K}{M}    
\end{gather}

Also, the spatial scaling term size of $(y_{demo,0}-g_{demo})$ has the same size as the original demo trajectory, thus the accumulated distance error between the demonstrated trajectory and the regenerated trajectory can be as the time integral of the Euclidean distance between the demonstrated trajectory interpolated to unit time and the regenerated trajectory for each time step. Therefore, it is concluded that the human-likeness term of the objective function should be
\begin{gather}
d(D_M, K_M) = \int_{0}^{1} ||y_{new}(D_M, K_M, t)-y_{demo}'(t)|| \,dt,\\
y_{demo}' = y_{demo}(T_{demo} t) - y_{demo}(0)
\end{gather}
where $d$ is the accumulated distance error as the human-likeness term of the objective function, $y_{demo}'$ is temporally normalized demonstrated trajectory with 1 sec duration and $T_{demmo}$ is the duration of the demonstrated trajectory. Due to the noise in acceleration and velocity, only the position profiles are compared in the human-likeness term.

Another important term to include in the objective function should relate to the stable representability of the demonstration trajectory. One of the primary assumptions of the DMP framework is that the similarity of the motions in the task space is conserved to the similarity of forcing terms (or the weights) in the DMP space~\cite{zhou2017taskdmp}. However, this assumption may be invalid when an improper set of dynamic features are chosen, where the DS cannot represent the motion robustly thereby causing unstable performance. In this sense, the similarity between the target forcing terms of multiple demonstrated trajectories can be measured over the parametric space of dynamic features $M,D,K$. While the DMP framework excels in one-shot learning with a single trajectory, analyzing multiple demonstrated trajectories can identify the overlaps of the human skills among these trajectories stably as a trade-off.

The similarity in the DMP framework between the trajectories can be defined as the standard deviation (SD) of the trajectories' forcing terms. For example, if the trajectories are ideally similar and all from a set of regenerated trajectories sharing the same forcing term with optimal dynamic features, the SD of the forcing terms of the trajectories will be zero. On the other hand, if the DS doesn't represent the original system because of an improper set of dynamic features, it may lose its capability to remain the similarity from the task space to the DMP space, thus the constructed forcing terms will be different in the DMP space even though the trajectories are similar in the task space.

Calculating the SD of the forcing terms in their original scale can cause an issue in the optimization. Since the SD becomes 0 where $M,D,K$ are all 0, it will only give a trivial solution as a global minimum. Instead, the topological similarity of the forcing terms can be considered by standardization of the forcing terms. In the same sense that $y_{demo}$ is temporally normalized in the accumulated distance error, $f$ can be temporally normalized by dividing the term by $M$, and spatially standardized as: 
\begin{equation}
\begin{aligned}
f_{target}(i,t) =& \ddot y_{demo}(i,t) + D_M\dot y_{demo}(i,t) \\
                 &+ K_M(y_{demo}(i,t)-g_{demo}(i)), \\
\end{aligned}
\end{equation}
\begin{align}
f_{stand}(i,t) =& \frac{f_{target}(i,t) - \overline {f_{target}(i)}}{\sigma(f_{target}(i,t))} 
\end{align}

\noindent
where $f_{target}(i,t)$, is the target forcing term from $i$th demonstration with a unit duration before the approximation with Gaussian basis, $f_{stand}(i,t)$ is the standardized target forcing term, $\overline {f_{target}(i,t)}$ and $\sigma(f_{target}(i,t))$ are the mean and the SD of the $i$th target forcing term over time. Then, the SD of $f_{stand,i}$ across the different demonstrations is calculated for each time step:

\begin{gather}
\overline{f_{stand}(t)} = \frac{1}{N} \sum_{i=1}^{N} f_{stand}(i,t), \\
\sigma_{f}(t) = \sqrt{\frac{1}{N-1} \sum_{i=1}^{N}{(f_{stand}(i,t) - \overline{f_{stand}(t)})^2}} 
\end{gather}

\noindent
where N is the number of the trajectories, $\overline{f_{stand}(t)}$ and $\sigma_{f}(t)$ are the mean and SD of $f_{stand}(i,t)$ across the trajectories at certain time step. As we have three DMP in x, y, z direction, $\sigma_{f_x}(t)$, $\sigma_{f_y}(t)$, $\sigma_{f_z}(t)$ can be obtained separately. Then, the final topological similarity term of the objective function can be achieved by taking the time integral of the norm of $\sigma_{f_x}(t)$, $\sigma_{f_y}(t)$, $\sigma_{f_z}(t)$:

\begin{gather}
S=\int \sqrt{\sigma_{f_x}(t)^2 + \sigma_{f_y}(t)^2 + \sigma_{f_z}(t)^2} dt
\end{gather}

\noindent
where $S$ is the topological similarity. $S$ is the function of $D_M$ and $K_M$. The suggested similarity function $S$ is designed to identify the optimal dynamic features where the forcing terms of the multiple demonstrated trajectories are similar.
 
In conclusion, the optimal solution of the dynamic features can be obtained by optimizing the objective function:
\begin{gather}\label{eq:obj}
J_{obj}(D_M,K_M)=S(D_M,K_M) + k\sum_{i=1}^{N}d_i(D_M,K_M)
\end{gather}
where $J_{obj}$ is the objective function, $S$ is the similarity between trajectories to check the stability of the DMP framework and $d_i$ is the $i$th accumulated distance error for each demonstration, $N$ is the number of the demonstration, $k$ is the adjustment gain for the objective function to balance the impact of each term. The result of the optimization gives $D$ and $K$ where $M$ is constant. 

\section{Experiment}

\subsection{Overview}

\begin{figure}[htbp] 
\centering
\includegraphics[width=1\linewidth]{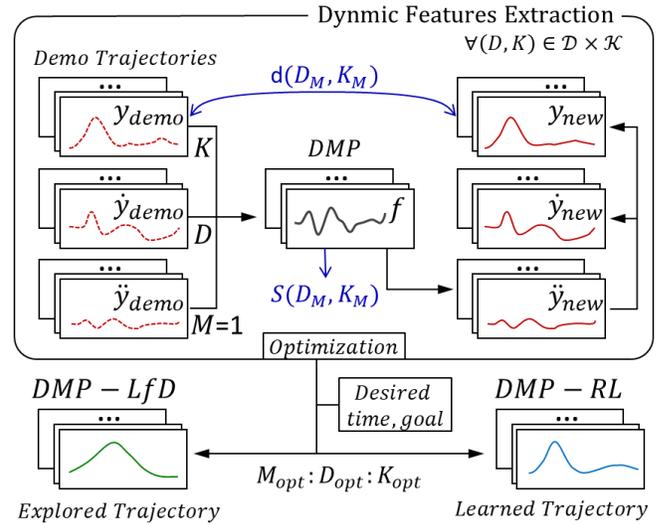}
\caption{Flow of dynamic feature extraction and implementation to LfD and RL with DMP. $M_{opt},D_{opt},K_{opt}$ are the extracted dynamic features achieved by the optimization of human-likeness term $d$ and motion stability term $S$.}
\label{fig:flow}
\end{figure}

The suggested skill transfer process, depicted in Fig. \ref{fig:flow}, involves multiple human demonstrations  to create the forcing terms in the DMP space. Dynamic features are optimized using the suggested methodology. After extraction, these features can be used for parameter tuning in the LfD with the DMP framework or applied to the RL with DMP framework to transfer human skills.

\subsection{Experimental setup}

The most common way to record human data would be using multiple vision cameras and the landmarks attached to the human skin, but it requires complicated setup and calibration. In this paper, we collected the trajectory in a simpler way by recording the position data of hand movements with HTC VIVE motion trackers 2018 and two SteamVR base stations. The 10 demonstrations were recorded by a human demonstrator for the point-to-point movement scenario.

When constructing the forcing term $f$ in the methodology, it is required to reduce the noise in the velocity and the acceleration in the demonstrated trajectories, as they are achieved from the time derivatives of position data. The Savitzky-Golay filter with the 3rd-order polynomial was applied, and the window length was set to 21. The number of the BFs used in DMP was 100.

\subsection{Feature Extraction}

\begin{figure}[htbp] 
\centering
\includegraphics[width=1\linewidth]{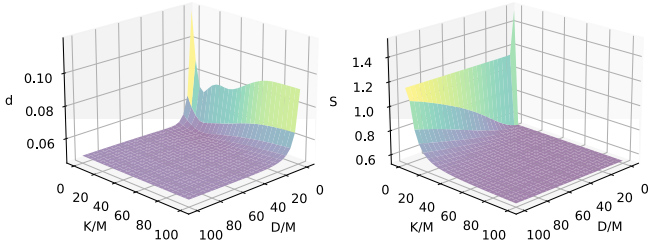}
\caption{The total accumulated distance error $\sum d$ of the demo trajectories (left) and their topological similarity $S$ (right)}
\label{fig:obj}
\end{figure}

To observe the contribution of $D_M$ and $K_M$ to the two terms in the objective function, the accumulated distance error and the topological similarity (Fig. \ref{fig:obj}) on the parametric space of $D_M$ and $K_M$ axes are plotted as below. The ranges of $D_M$ and $K_M$ are bounded to be positive.

As shown in the accumulated distance error, too small spring and damping coefficients $D$ and $K$ compared to the inertia $M$ of a DS hinder the ability to regenerate the motion close to the original demonstration. In contrast, the similarity map of $S$ shows that when $K$ is too smaller than $M$ in the DS, the DMP cannot stably represent the common skill in the demonstrated trajectories even though they are similar in the task space. Also, it was observed that the higher $D_M$ and $K_M$ give slightly higher $d$ and $S$ in the plain region of the graph. Thus, the objective function including two terms can find the balance between the two goals represented by two terms of the objective function. The adjustment gain $k$ introduced in the objective function (Eq.~\eqref{eq:obj}) was set to 20, and the extracted ratio of the dynamic features was
\begin{align}
(M:D:K)_{extracted} = 1.00 : 10.73 : 20.71
\end{align}
with the damping ratio $\zeta=1.28$, resulting in an over-damped DS in the DMP.

\subsection{DMP Implementation}

In the LfD with DMP setup, one of the demonstrated trajectories was picked to learn its motion using the extracted dynamic features. The time duration of the regenerated trajectory was set to 1 sec to compare with the temporally normalized demo trajectory. Also, the goal position remained the same as the original goal position.

In the RL with DMP setup, the agent's action substituted the forcing term. The action was used to calculate the next acceleration, velocity and position of the trajectory. The sparse rewards were given to the agent. During the episode, the reward was given as
\begin{align}
r = -10^{-3} |\ddot y| - 10^{-7}|A|,
\end{align} 
and at the end of the episode,
\begin{align}
r = -10*|y-g_{new}|
\end{align} 
where $A$ is an action, $g_{new}$ is the goal position of the new trajectory that was the same as $g_{demo}$. The time duration of the new trajectory was set to 1 sec. Proximal Policy Optimization (PPO) algorithm was used with the hidden size (64,64) for both actor and critic layers.

\section{Results}

We applied the extracted dynamic features to regenerate trajectories using LfD with DMP, with the same goal position as the demonstrated trajectory. We also used the heuristic $D_M, K_M$ in Table \ref{tab:comparison} to regenerate other trajectories for comparison. Demo 1 and 2 are the demonstrated trajectories used to learn DMP.

\begin{figure}[htbp] 
\centering
\includegraphics[width=0.9\linewidth]{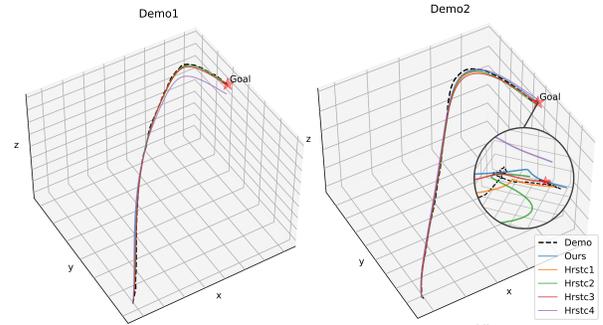}
\caption{The path of regenerated trajectories using the extracted dynamic feature and the heuristic features in the LfD with DMP framework.}
\label{fig:demo_path}
\end{figure}

\begin{figure}[htbp] 
\centering
\includegraphics[width=0.9\linewidth]{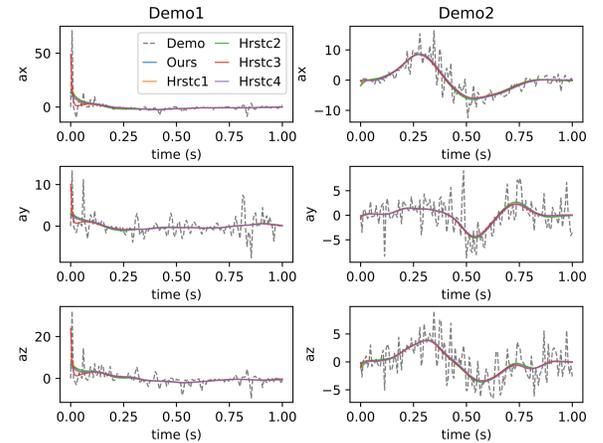}
\caption{The accelerations of regenerated trajectories using the extracted dynamic feature and the heuristic features of the LfD with DMP framework.}
\label{fig:demo_acc}
\end{figure}

\renewcommand{\arraystretch}{1.15}
\begin{table}[htbp]
\caption{Performance comparison of the extracted (Ours) and the heuristic (Hrstc1 to Hrstc4) dynamic features}
\label{tab:comparison}
\resizebox{\linewidth}{!}{
\begin{tabular}{lrrr|rc|rc|}
\cline{5-8} & \multicolumn{1}{c}{} & \multicolumn{1}{c}{}      & \multicolumn{1}{c|}{}       & \multicolumn{2}{c|}{\textbf{Demo1}}                    & \multicolumn{2}{c|}{\textbf{Demo2}}         \\ \hline
\multicolumn{1}{|l|}{\textbf{Method}} & \multicolumn{1}{c|}{\textbf{D/M}} & \multicolumn{1}{c|}{\textbf{K/M}} & \multicolumn{1}{c|}{$\zeta$} & \multicolumn{1}{c|}{$d_{mean}$} & $a_{peak}$ & \multicolumn{1}{c|}{$d_{mean}$} & $a_{peak}$ \\ \hline
\multicolumn{1}{|c|}{\textbf{Ours}}  & \multicolumn{1}{r|}{10.73}  & \multicolumn{1}{r|}{20.71}  & 1.18  & \multicolumn{1}{r|}{3.58mm} & 13.35m/s$^{2}$ & \multicolumn{1}{r|}{7.54mm} & 8.64m/s$^{2}$ \\ \hline
\multicolumn{1}{|c|}{\textbf{Hrstc1}} & \multicolumn{1}{r|}{25.00}  & \multicolumn{1}{r|}{156.25} & 1.00  & \multicolumn{1}{r|}{6.24mm} & 19.69m/s$^{2}$ & \multicolumn{1}{r|}{7.84mm} & 8.64m/s$^{2}$ \\ \hline
\multicolumn{1}{|c|}{\textbf{Hrstc2}} & \multicolumn{1}{r|}{10.00}  & \multicolumn{1}{r|}{200.00} & 0.35  & \multicolumn{1}{r|}{6.88mm} & 13.91m/s$^{2}$ & \multicolumn{1}{r|}{8.75mm} & 8.68m/s$^{2}$ \\ \hline
\multicolumn{1}{|c|}{\textbf{Hrstc3}} & \multicolumn{1}{r|}{100.00} & \multicolumn{1}{r|}{20.00}  & 11.18 & \multicolumn{1}{r|}{5.72mm} & 48.95m/s$^{2}$ & \multicolumn{1}{r|}{7.61mm} & 8.44m/s$^{2}$ \\ \hline
\multicolumn{1}{|c|}{\textbf{Hrstc4}} & \multicolumn{1}{r|}{4.00}    & \multicolumn{1}{r|}{4.00}   & 1.00  & \multicolumn{1}{r|}{16.82mm} & 10.59m/s$^{2}$ & \multicolumn{1}{r|}{7.51mm} & 8.51m/s$^{2}$ \\ \hline
\end{tabular}
}
\end{table}
Fig.\ref{fig:demo_path} and Fig.\ref{fig:demo_acc} display the regenerated trajectories, learned from Demo 1 and Demo 2 Both in Demo 1 and Demo 2, the regenerated trajectories made with the extracted dynamic features followed the demonstrated trajectories as well as the fine-tuned Hueristic1. In contrast, Heuristic2 with the high $K_M$ and the 
 low damping ratio $\zeta$ showed too much reactive motion to the small pause of the demonstration near goal. Hueristic3 with the high $D_M$ and the  high $\zeta$ was vulnerable to the acceleration noise in the original trajectory showing the initial overshoot in its acceleration. Heuristic4 failed to reach to the goal position even though $\zeta$ was selected to 1 following the suggestion in ~\cite{ijspeert2013dynamical}. Table~\ref{tab:comparison} shows the mean distance error $d_{mean}$ which is the accumulated distance error $d_{mean}$ divided by the time duration of the motion, and the peak acceleration $a_{peak}$ of each generated trajectory.

\begin{figure}[htbp] 
\centering
\includegraphics[width=0.75\linewidth]{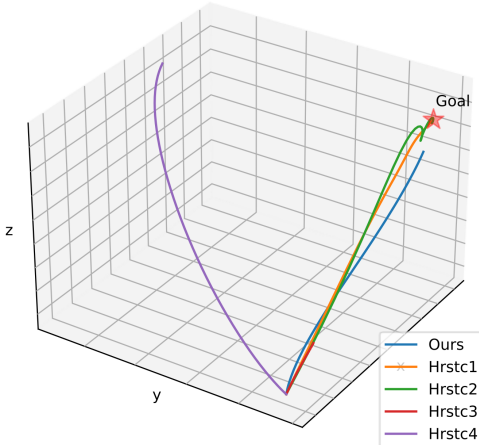}
\caption{The path of regenerated trajectories using the extracted dynamic feature and the heuristic features in the RL with DMP framework.}
\label{fig:rl}
\end{figure}

Also, the dynamic features were applied to RL with DMP. Fig.\ref{fig:rl} shows that trajectory regeneration is more sensitive to the dynamic feature selection compared to LfD with DMP. While both the extracted dynamic features and fine-tuned Heuristic1 were able to reach the goal position robustly, the regenerated trajectories using other heuristic features either failed to reach the goal position entirely or produced ineffective trajectories.

\section{Discussion}

The extracted features generated a trajectory that performed as robustly and stably as the fine-tuned heuristic features, while the trajectories generated with the heuristic features before tuning resulted in sudden acceleration or failure to reach the goal. The RL with DMP framework was more sensitive to the right choice of dynamic features. The extracted features showed stable performance as well as LfD with DMP, delivering the dynamic characteristic extracted from human demonstration.

Further investigations could explore different schemes to improve the objective function. For example, the time integral used for Euclidean distance and standard deviation may emphasize distance and similarity when motions are misaligned in time. The methods invariant to time shifts, such as phase correlation, could be used to overcome this limitation.

\section{Conclusions}

In this paper, we proposed an original methodology to extract dynamic features from the demonstrated trajectories which can be used in the DMP framework. The DMP using the extracted values performed as well as the DMP using carefully tuned heuristic parameters. Hence, the proposed method not only omits the time-consuming manual parameter tuning but also increases the robot's capability to learn the motion autonomously. Furthermore, the extracted values can be used for the DMP framework in conjunction with RL. In this way, the robot can explore more potential trajectories and find the optimal one instead of being restricted to the demonstrated trajectory. It is worth emphasizing the result of extracting human dynamic features is enhanced compliance of the robot with human intentions.
In conclusion, the proposed dynamic feature extraction in the DMP framework bridges the LfD and RL, thereby delivering a human skill besides the trajectory to the robot.
The key idea of dynamic feature extraction is to maximize the human-likeness while guaranteeing the representability of its dynamic system across similar motions in the task space. The implementation of the extracted value into the real robot showed that the robots reproduce trajectories robustly and stably throughout the motion.
There are several areas for further investigation. While DMP uses second-order dynamics, the proposed dynamic feature extraction can be examined for higher-order system dynamics (e.g., jerk in a third-order equation of motion). Also, further research may investigate the application of the suggested framework to other variations of DMP designed for different scenarios, such as learning obstacle avoidance or kicking motion.

\addtolength{\textheight}{-12cm}   








\bibliographystyle{./IEEEtran}
\bibliography{./IEEEabrv,./reference}

\end{document}